\documentclass[conference]{IEEEtran}
\IEEEoverridecommandlockouts

\usepackage{latexsym}
\usepackage{multirow}
\usepackage{lipsum}
\usepackage{graphicx}
\usepackage{url}
\usepackage{float}
\restylefloat{table}
\usepackage{placeins}
\usepackage[bottom]{footmisc}
\newcommand*{\thead}[1]{\multicolumn{1}{|c|}{\bfseries #1}}

\usepackage{cite}
\usepackage{amsmath,amssymb,amsfonts}
\usepackage{algorithmic}
\usepackage{graphicx}
\usepackage{textcomp}
\usepackage{xcolor}
\usepackage{multirow}
\usepackage{array}
\usepackage{multicol}
\usepackage{etoolbox}
\usepackage{bm}
\usepackage{subfig}
\usepackage{etoolbox}
\makeatletter
\patchcmd{\@makecaption}
  {\scshape}
  {}
  {}
  {}
\makeatother

\def\BibTeX{{\rm B\kern-.05em{\sc i\kern-.025em b}\kern-.08em
    T\kern-.1667em\lower.7ex\hbox{E}\kern-.125emX}}
\begin{document}

\title{A Novel Neural Sequence Model with Multiple Attentions for Word Sense Disambiguation \\
{\footnotesize \textsuperscript{}}
}

\author{\IEEEauthorblockN{Mahtab Ahmed, Muhammad Rifayat Samee, Robert E. Mercer}
\IEEEauthorblockA{\textit{Department of Computer Science} \\
\textit{University of Western Ontario}\\
London, Ontario, Canada \\
mahme255, msamee, rmercer@uwo.ca}}

\maketitle

\begin{abstract}
  Word sense disambiguation (WSD) is a well researched problem in computational linguistics. Different research works have approached this problem in different ways. Some state of the art results that have been achieved for this problem are by supervised models in terms of accuracy, but they often fall behind flexible knowledge-based solutions which use engineered features as well as human annotators to disambiguate every target word. This work focuses on bridging this gap using neural sequence models incorporating the well-known attention mechanism. The main gist of our work is to combine multiple attentions on different linguistic features through weights and to provide a unified framework for doing this. This weighted attention allows the model to easily disambiguate the sense of an ambiguous word by attending over a suitable portion of a sentence. Our extensive experiments show that multiple attention enables a more versatile encoder-decoder model leading to state of the art results.
\end{abstract}

\section{Introduction}

Word sense disambiguation (WSD) is the task of assigning appropriate meaning to a target word in the case where the target sense is clearly distinguishable from other word senses subjected to the attributes of the target word’s context. As one of the challenging problems in the field of computational linguistics, WSD has received considerable attention over the past decade  \cite{agirre2007word, navigli2009word} due to its various application potentials such as information retrieval, text mining, machine translation, speech synthesis as well as question answering etc. 
Some of the classical algorithms for solving WSD are LESK algorithm which sees the dictionary definition overlapping of the words in a sentence, Naive Bayes; which looks at the conditional probability of each word sense along with the contexts having an assumption that the ordering of words as well as their presence among a bag of words is independent and Neural Networks; where the words in a sentence are represented by nodes and they gradually turned on / off as the training goes on; at the same time activating neighbour nodes on the next cycle and finally stabilizes in a state where one sense for each input word is more activated than the others.

A large amount of research works are continuously going on to solve this classical WSD problem, developing new algorithms \cite{yuan2016semi, butnaru2017shotgunwsd, tripodi2017game} and evaluating on some of the famous benchmarks \cite{snyder2004english, navigli2007semeval, moro2015semeval}. All of these recent works mainly focus on the two famous WSD drawbacks; order of the words in the context and series of handcrafted features. Most of the traditional supervised WSD methods are based on extracting features form the surrounding words and then train a classifier for each of the ambiguous words \cite{zhong2010makes}.

Recently, Deep neural network (DNN) based approaches have gained state of the art results in many widely examined classical problems in computational linguistic field. In the past few years, neural network has been very successfully applied for getting the word embedding; representing each word as a vector of fixed dimensions. Mikolov et al. \cite{mikolov2013distributed} trained a shallow neural network to get vector representation of a target word depending on its surrounding contexts and claims that their representation captures the syntactic and semantic level similarity very well. Later on, Levy et al.\cite{levy2014neural} proved that Mikolov's word2vec skip-gram model with negative sampling is actually an implicit factorization of a word context matrix in which every cell is the point-wise mutual information (PMI) of the respective word and context pairs shifted by a global constant.

Neural machine translation (NMT) is a new way to deal with machine translation, recently proposed by \cite{kalchbrenner2013recurrent,sutskever2014sequence, bahdanau2014neural}. Unlike the conventional phrase-based translation framework \cite{koehn2003statistical} which comprises of numerous small sub-segments that are tuned independently, NMT endeavors to construct and prepare encoder-decoder based vast neural network that goes through a sentence and yields a right translation. The attention mechanism has been a breakthrough in NMT which calculates How much attention the network has to give on each source word to generate a specific translated word. Bahdanau et al. \cite{bahdanau2014neural} first proposed this concept of doing translation as well as alignment jointly, where they calculate the context using encoder outputs and last hidden state of decoder at each time step. Finally, they combined this context with the decoded word from previous time step to generate a new word in current time step. Luong et al. \cite{luong2015effective} came up with two new attention mechanisms where one looks for the global context and the other one looks for local context i.e. a subset of words. They used translated word in each time step along with encoder outputs to calculate the context. Finally, they concatenated this context with the recurrent neural network (RNN) output of the decoder and mapped it to a translated word via a multi layer perceptron (MLP). Sennrich et al. \cite{sennrich2016linguistic} show that, although these Seq2Seq models captures word level features very well, it would not be redundant to add some linguistic features such as part-of speech tags, morphological features as well as syntactic dependency and they found improved results on WMT16 task.

A number of recent works adopted this sequence to sequence (Seq2Seq) concept for doing WSD. Among them, Raganato et al. \cite{raganato2017neural} have experimented with some neural sequence models which include bidirectional long short term memory (BLSTM) based architecture in many to many form with and without attention for sequence tagging. They also experimented on sequence to sequence architecture with attention to do multitask learning where they tag a sentence with their sense as well as their parts of speech. Their best performing model is attentive BLSTM tagger rather than the Seq2Seq architecture. Melamud et al. \cite{melamud2016context2vec} trained a BLSTM architecture to get the context representation of each sense annotation on an unlabeled corpus. They used an objective function where the sentence with the target word is empty or replaced by a dummy symbol and then try to put that modified context as well as the target word in the same low-dimensional space. Kaageback et al. \cite{kaageback2016word} relied on a BLSTM based approach where they divided the sentence in two sections based on the position of the target words. They call it left context and right context. Finally, they applied two long short term memories (LSTMs) from two opposite directions on these contexts, concatenated their last hidden states and used an MLP to classify to corresponding sense. Yuan et al. \cite{yuan2016semi} propose a powerful neural language model to obtain a latent representation for the entire sentence containing a target word \textit{w}; later on, they compare this representation with those sentences which have other candidate meanings of word \textit{w}. Ahmad et al. \cite{Ahmad18} came up with an architecture, where they calculate the cosine similarities between the sense embedding of the center word and the word embedding of every other words in a sentence. Then they applied two LSTMs on this vector of similarities from both left and right directions, concatenated them and finally applied a fully connected layer to classify the word sense as a one class classification problem. 

In this paper, we propose a few encoder-decoder architectures for doing WSD by taking linguistic features of the surrounding context words into account, combining them and at the same time capturing the order of the context words as well. The Seq2Seq architecture for NMT is being used in this work as a sequence tagger and the attention mechanism is used to tell the network, how much attention it needs to pay on each linguistic feature to identify the specific meaning of an ambiguous word. We choose neural sequence models because it generally perform very well on sequence data by keeping them in the memory \cite{raganato2017neural} and adding linguistic features with these models actually improve their performance \cite{sennrich2016linguistic}.
Although Raganato et al. \cite{raganato2017neural} show that Seq2Seq is sub-optimal for doing WSD, this paper revisit their finding and explore the effectiveness of various attentive encoder-decoder architecture for this task. We are using supervised attention-based method to come up with a linear combination of different features and generate a final linguistic feature based attention matrix. As candidate features, we are looking at three possible ones: surrounding word vectors, surrounding context bigrams and the parts of speech (POS) sequence of the whole sentence. Adding multiple attentions help to decide the significant contribution of different attentional features and a vector representing linear combination of those attentional features captures every pieces of it. Even though we haven't been able to take the entire corpus into account because of the resource limitation, still it’s been found that our Seq2Seq architecture with multiple attentions have obtained state of the art performance on some of the benchmarks. 

\begin{figure*}[ht]
  \includegraphics[width=\textwidth,height=8.15 cm]{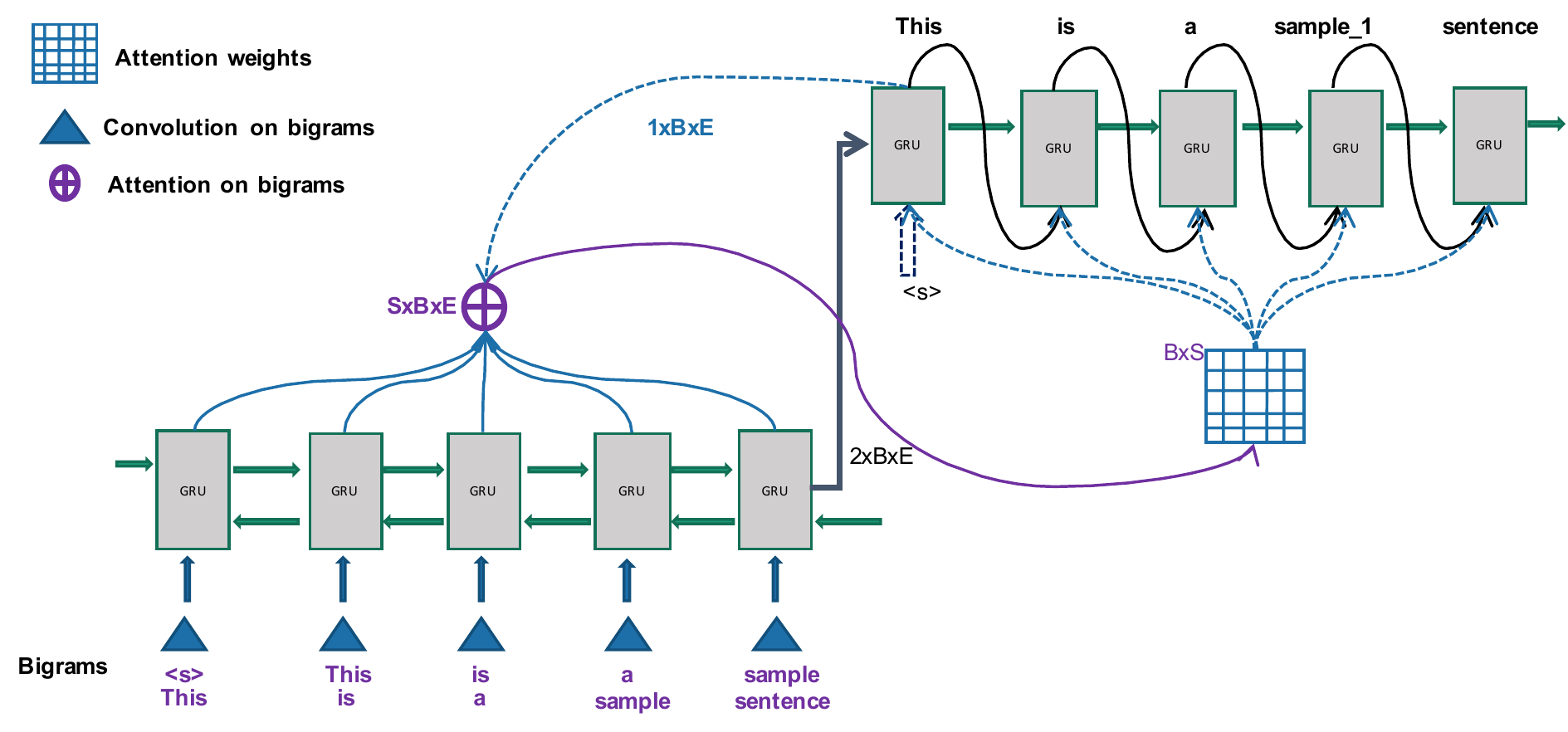}
  \caption{\label{bigram}Encoder-decoder architecture for sequence-to-sequence WSD with attention on bigrams.  (\textbf{S} =  Sentence length, \textbf{B} = Batch size and \textbf{E} = Embedding dimension)}
\end{figure*}
\section{The model}

In this section, we describe our work in detail. We first explain our basic Seq2Seq model having attention on bi-grams. Following this, we explain a way of doing multiple attentions on different features and finally, we describe a way of combining these multiple attentions using some weighted value in the latter part of this section.
\begin{figure*}[ht]
  \includegraphics[width=\textwidth,height=8.15 cm]{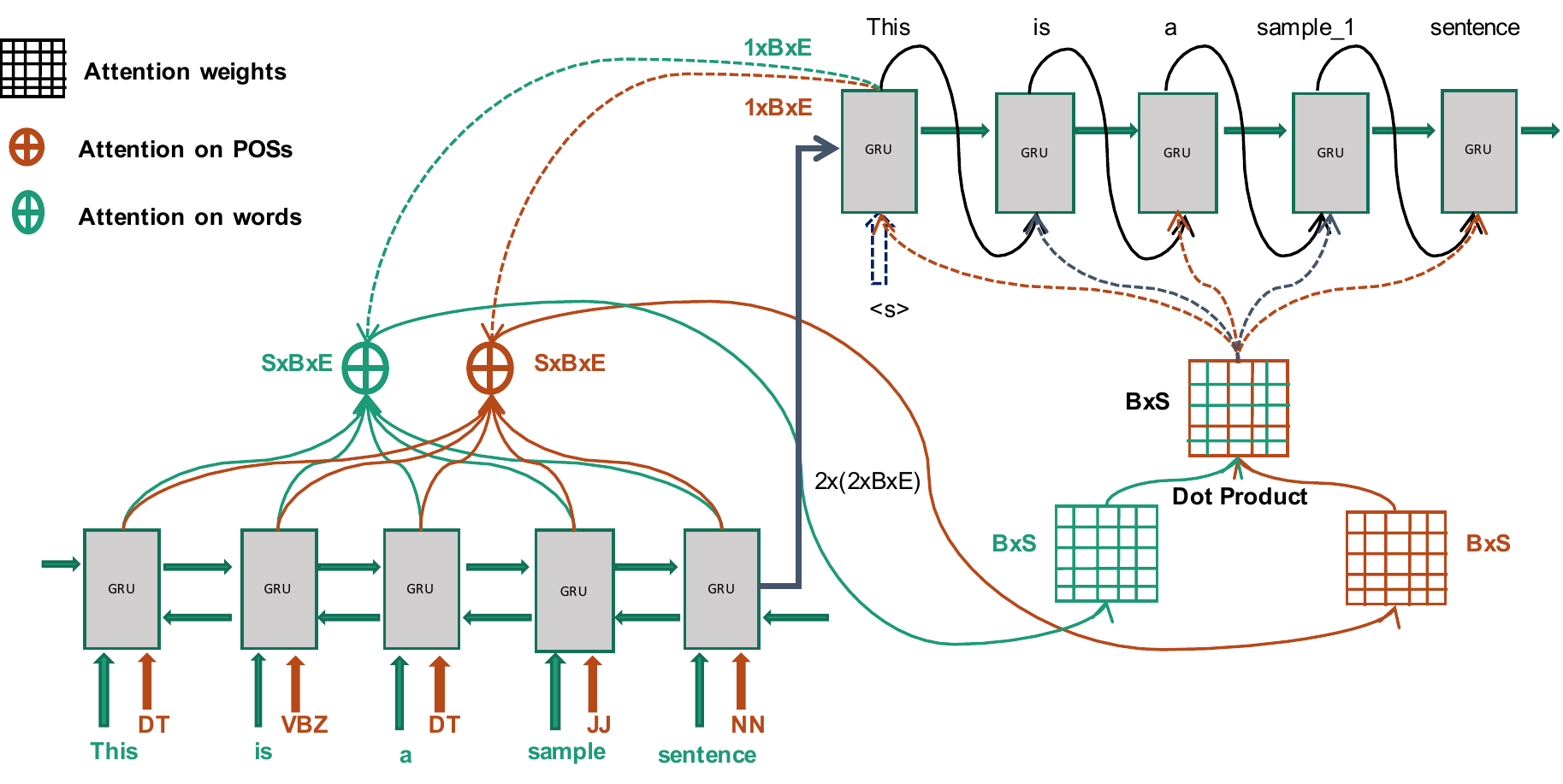}
  \caption{\label{POS} Encoder-decoder architecture for sequence-to-sequence WSD, taking point-wise multiplication of POS and bigram attention weights.  (\textbf{S} =  Sentence length, \textbf{B} = Batch size and \textbf{E} = Embedding dimension)}
\end{figure*}
\subsection{Attention on Bigrams}\label{model1}
Our first architecture is completely based on the attention based Seq2Seq model with encoders and decoders as shown in Figure \ref{bigram}. The encoder inputs the source sentence as a sequence $[x_1,x_2, \ldots,x_t,\ldots, x_n]$ , having $x_t$ as the ambiguous word. The decoder tries to generate a sequence $[y_1,y_2,\ldots,y_t,\ldots,y_n]$  where $x_i = y_i$ for all \textit{i}, except the target word at index \textit{t} is replaced by the corresponding sense tagged word. Adding attention mechanism with this architecture allows the model to take one word at a time from the decoder and looks for which of the words in the input sentence are useful for generating this target word. However, rather than using word by word attention, we use bigram attentions which actually makes sense, because to generate a particular sense of a word, the context words contribution matters most. Next, we pad a start token $\textless s\textgreater$ at the beginning of this sentence and then pass this modified sentence to an embedding layer. The embedding weights are initialized randomly as well as using pre-trained word vector and are trained along with other parameters of the network. Next, a convolution layer with kernel size $2 \times embedDim$ is initialized which goes over the bigram embedding with a stride length of 1 as shown in Eqn. \ref{first-equ}.

\begin{equation}  \label{first-equ}
B_{i,:}=\ \sum^2_{j=1}{E_{i+j,:}*K_{j,:}} 
\end{equation}  
where, $E$ is the embedding matrix, $K$ is the convolution kernel and $n$ is the maximum sequence length for the current batch. This will generate a convolved embedding of bigrams $B$ and it is then fed to the gated recurrent unit (GRU) layer.
\begin{equation} \label{encoder}
 h_t = \texttt{GRU}(B_t,h_{t-1})
\end{equation}
The last hidden state $h_n$ is the encoded representation of the sentence and we term this as $h_{enc}$. Next, in the decoder section, a GRU is initialized with $h_{enc}$ as the hidden state and $\textless s\textgreater$ as input. This generates a new hidden state $\tilde{h}_t$.
\begin{equation}\label{decode}
    \tilde{h}_t = \texttt{GRU}(\textless s\textgreater, h_{enc})
\end{equation}
 We then pass this $\tilde{h}_t$ and encoder output $O$ at each time step to an `Attention' model which returns an attention matrix $A$ of size $1 \times n$.
 \begin{equation}\label{attention}
     A = \texttt{Attention}(O, \tilde{h}_t)
 \end{equation}
Next, we apply batch-wise matrix multiplication on $A$ and $O$ and compute the context $C_{1 \times d}$.
\begin{equation}\label{context}
    C = A.bmm(O)
\end{equation}
We then concatenate $\tilde{h}_t$ and $C$ and pass it to an MLP followed by a \texttt{Softmax} layer which maps the result back to the vocabulary size
\begin{equation}
\begin{split}\label{softmax}
    \tilde{Y}_t = \texttt{MLP}([\tilde{h}_t, C])\\
    Y_t = \texttt{Softmax}(\tilde{Y}_t)
    \end{split}
\end{equation}
In the next time step, the decoder GRU again unfolds. But this times it takes hidden state and the word generated from previous time step into account. So, Eqn. \ref{decode} becomes,
\begin{equation}\label{decoder}
    \tilde{h}_t = \texttt{GRU}(\tilde{Y}_{t-1}, \tilde{h}_{t-1})
\end{equation}
The rest of the training is done as an end to end fashion.
\begin{figure*}[ht]
\includegraphics[width=\textwidth,height=8.15 cm]{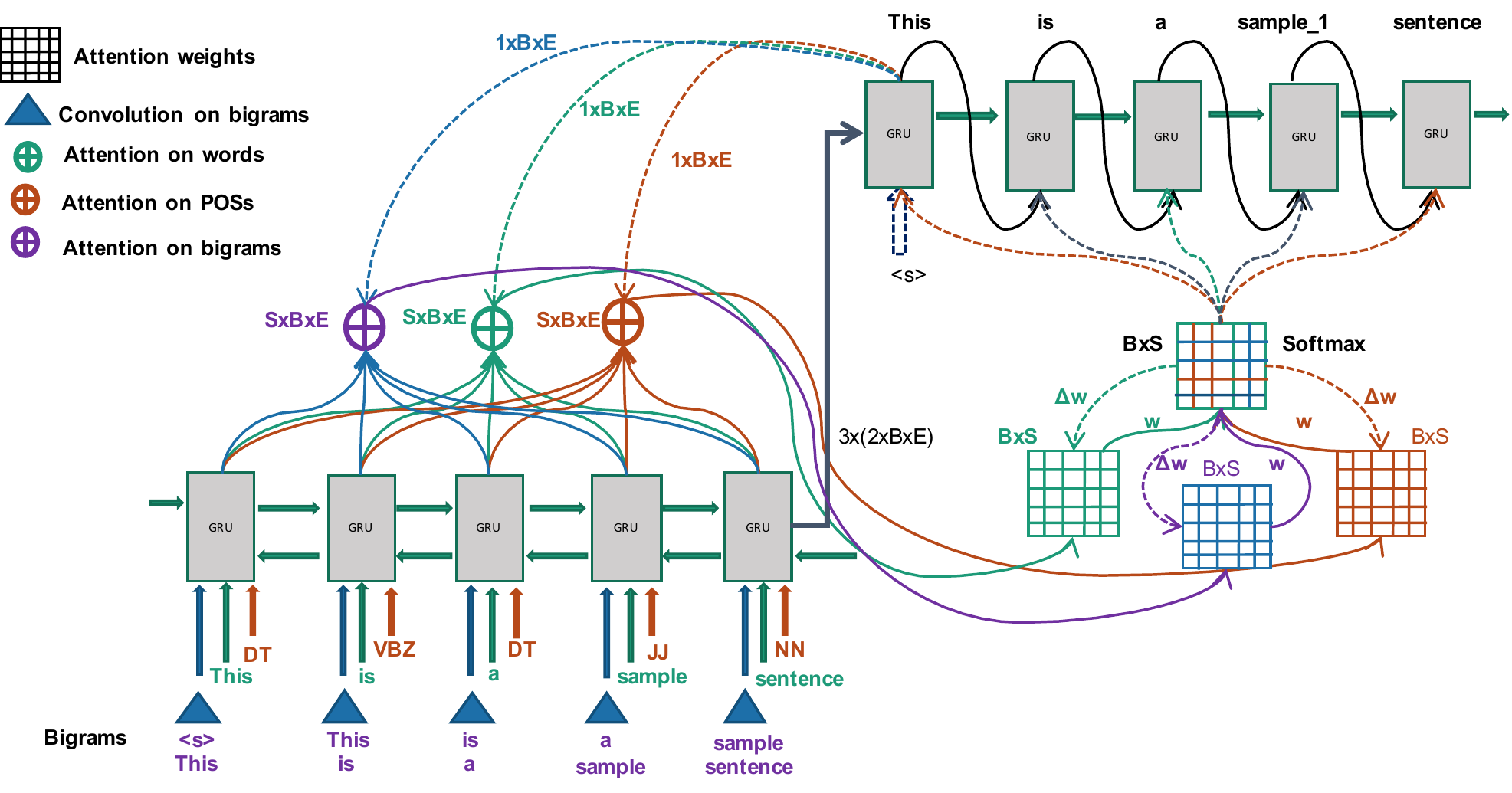}
  
\caption{\label{bigram-POS}Encoder-decoder architecture for sequence-to-sequence WSD with weighted combination of multiple attentions. (\textbf{S} =  Sentence length, \textbf{B} = Batch size and \textbf{E} = Embedding dimension)}
\end{figure*}

\subsection{Attention on Words and Parts of Speech (POS)}\label{model2}

In this model, we introduce multiple attention mechanism where we apply individual attention on different features of data and later combine them through point-wise multiplication. We start with traditional Seq2Seq architecture having encoder and decoder at both ends. As shown in Figure \ref{POS}, the encoder has GRU Layer which takes word embeddings as well as POS embeddings as input. Rather than having different GRU layer for words and POS, we use a single GRU layer whose weights are shared for both of these inputs. This also allows the gradients to be shared as well during back-propagation. Next, using Eqn. \ref{encoder}, we calculate the encoded word and POS features as follows,
\begin{equation}
    \begin{split}
        h^{w}_{t} = \texttt{GRU}(W_t, h_{t-1})\\
        h^{p}_{t} = \texttt{GRU}(P_t, h_{t-1})
    \end{split}
\end{equation}
where $h^{w}$ and $h^{p}$ represent the hidden state for word and POS respectively. From each of these hidden state vectors, we consider the one at the last index as the encoded version. Following this, we take these encoded hidden state vectors and apply Eqn. \ref{decode} and \ref{attention} which gives us two attention matrices $A^w$ and $A^p$ for word and POS respectively. We get the final attention matrix $\tilde{A}$ by doing a point-wise multiplication of $A^w$ and $A^p$,
\begin{equation}\label{second-equ}  
    \tilde{A}= A^w *  A^p
\end{equation}  
The point-wise multiplication changes the amplitude of each dimension of these vectors and also allows encoding the positions of the target word neighbors according to their vector amplitude. If both word and POS attention matrix puts more focus on $i$th word then the magnitude of the final attention vector at $i$th dimension becomes very large. And if two attention vectors put focus on two different words, then the final attention gets distributed over those two possible words. This makes it similar to a soft-hard attention model, where the model decides which one gets activated and when. Next, to make the final attention matrix $A$ as a probability distribution, we apply a \texttt{Softmax} layer on it 
\begin{equation}\label{final_softmax}
    A = \texttt{Softmax}(\tilde{A})
\end{equation}
Finally, we use Eqn. \ref{context} to calculate the context followed by Eqn. \ref{softmax} to generate the next predicted word. Similarly to the previous subsection \ref{model1}, Eqn. \ref{decode} gets replaced by Eqn. \ref{decoder} from the second time step to generate the hidden state vector for decoder. The decoder continues to decode until it generates all words in the target sentence or an EOS token is encountered. The rest of the training is done in end to end manner.

\subsection{Attention on Words, Parts of Speech and Bigrams with weighting} \label{baseline_weighting}
In this model, we combine the concept of both bigrams and POS attention architecture from \ref{model1} and \ref{model2} and introduces a term called weighted attention. As shown in Figure \ref{bigram-POS}, input GRU in the encoder takes word embeddings, POS embeddings as well as the bigram embeddings. Similarly to our previous model, we share the GRU weights among these three inputs. We use Eqn. \ref{encoder} on these three inputs independently and calculate three encoded vectors $h^w$, $h^p$ and $h^b$ for word, POS and bigram respectively. Following this, we apply Eqn. \ref{decode} and Eqn. \ref{attention} on these encoded vectors independently and calculate three attention vectors $A^w$, $A^p$ and $A^b$. Next, we generate three weights $w_1$, $w_2$ and $w_3$ and perform a weighted linear addition of three attention vectors using them, 
\begin{equation}
    A = (A^w * w_1) + (A^p * w_2) + (A^b * w_3)
\end{equation}
Following this, we use Eqn. \ref{final_softmax} to turn these attention weights into probability. Finally, calculating the context and generating the next probable word is similar to the one in \ref{model1} and \ref{model2}. The decoder continues to generate until an EOS token is found or the entire sequence gets generated.

\textbf{Generating \textbf{weight} values} Currently, most of the research works use attention to decide which portion of a feature the model needs to attend to more. To the best of our knowledge, no work has investigated putting attention on attentions. In this study, we propose a few ways to do this. Firstly, if a model has multiple features with separate attention for each, the simplest method is to combine all the attentions and perform a \texttt{Softmax} on them. Another way is to apply a local gating mechanism, where we first pass each of the individual attention vectors to a \texttt{Sigmoid} layer. This generates a value in $[0,1]$ for each attention vector. Then we perform a point-wise multiplication on these individual attention vectors and their sigmoid values. Finally, we add up these values, apply a \texttt{Softmax} on them and use the result as the attention vector. 
\begin{equation}
    \begin{gathered}
        \tilde{A}_1 = \texttt{Sigmoid}(A_1)\\
        \tilde{A}_2 = \texttt{Sigmoid}(A_2)\\
        \tilde{A}_3 = \texttt{Sigmoid}(A_3)\\
        \tilde {A} = (\tilde{A}_1 * A_1) + (\tilde{A}_2 * A_2) + (\tilde{A}_3 * A_3)\\
        A = \texttt{Softmax}({\tilde{A}})
    \end{gathered}
\end{equation}
Apart from local gating, it is also possible to apply a global gating mechanism where we first concatenate all the attention vectors and separately apply \texttt{Softmax} on individual columns of this concatenated matrix. We then pick the best features for each position via \texttt{argmax} and make a vector from it. Finally we apply another \texttt{Softmax} on this vector to get the attention vector.  
\begin{equation}
    \begin{gathered}
        \hat{A} = [\hat{A}_1; \hat{A}_2; \hat{A}_3]\\
        \tilde{A}_{:,i} = \texttt{Softmax}(\hat{A}_{:,i}), \forall i \\
        \tilde{A} = [\tilde{A}_{:,1};\tilde{A}_{:,2};\tilde{A}_{:,3}]\\
        A = \texttt{Softmax}({\tilde{A}})
    \end{gathered}
\end{equation}

Another way to generate the weights is to scale each of the attention vectors with a scaling factor and then add them. The factor can be a vector or a scalar. To do this, first we initialize the factor with some random values and then add it to the model parameters where its gradient is calculated based on loss. If the factor is a vector then we can choose an MLP without a bias for doing the transformation. In this study, we choose scalar factors as weights, initialize them to $1.0$ and then perform a linear weighted addition of the attention vectors. After the addition, we apply a \texttt{Softmax} on the resultant vector in order to make it a vector of probabilities. In the next section we show that even though we start with  weight values of $1.0$, by taking gradients during training, the model adjusts the values accordingly and we end up with a completely different set of values. 

\section{Experimental Setup}
In this section we describe the detailed experimental setup for the evaluation of our study. We first explain our training corpus as well as all the benchmarks used in other standard WSD methods. Following this, we explain the technical details of our proposed architectures along with their hyper-parameter settings.

\textbf{Training and Evaluation Benchmarks:} We use SemCor 3.0, MASC and Senseval task 3 corpora for training as well as evaluating our models. Many existing works have used these corpora as their benchmark \cite{yuan2016semi, kaageback2016word, taghipour2015semi,grozea2004finding, strapparava2004pattern, lee2004supervised}.
As our models are configured to work with WordNet senses, we use the mapping algorithm proposed by \cite{yuan2016semi} to map the SemCor and MASC corpora from NOAD senses to WordNet senses. Moreover, we also evaluate our architectures both on the same domain as well as on the different domains. During this cross domain evaluation, we train on one corpus and test on another. Standard splits of these corpora are used during evaluating on the same domain (Train - SemCor, Test - SemCor and vice versa). And during the cross domain evaluation, the whole corpus from one domain is used for training and the entire corpus from other domain is used for testing.

\begin{table}[b!]
\centering
\small
\begin{tabular}{p{1.5in}p{1.2in}} \hline 
\textbf{Hyper-parameter} & \textbf{Range Selected} \\ \hline 
Learning rate & 0.01 / 0.02 / 0.001 \\ \hline 
Context size & 50 (25 on each side) \\ \hline 
Batch size & 10 / 50 / 100 \\ \hline 
No. of GRU layers  & Encoder- 2 \newline Decoder - 2 \\ \hline 
Type of GRU layer  & Encoder- bidirectional \newline Decoder- unidirectional \\ \hline 
Dropout & 0.1 / 0.2 / 0.3 \\ \hline 
Word embedding size & 100 / 200 / 300 \\ \hline 
Initialization of scalar weights on Attentions & Random $\in$ uniform \newline (-0.1, 0.1) \\ \hline 
Decoder learning ratio & 5.0 \\ \hline 
Gradient clipping & 50 / 25 / 10 \\ \hline
\end{tabular}
\caption{\label{hyper}  Hyper-parameter used for the experiments and the ranges that were searched during tuning. }
\end{table}

Model selection is done using the validation set of each corpus and the best model is finally trained by combining the training and the validation set to evaluate on the left alone test set. As we have not found any standard splits for SemCor and MASC corpora, we perform a manual split: train, test and validation with a ratio of 80\%, 10\%, and 10\% respectively.
The hyper-parameters are tuned solely according to the validation set.

During testing, our models calculate the probability distribution over output words $O$ given a target word $w$. The output $O$ at each time step is fed to a \texttt{Softmax} layer  which gives the probability for each class. It is then used to rank the candidate senses of $w$ and the top ranked candidates are selected as the output of the model. 

\textbf{Architecture details and network parameters:} For all three architectures, we use GRU as the basic building block. Only for the first architecture, we use single attention and for the other two architectures, we use multiple attentions. We use the `dot' , `concat' and `linear' attention model from \cite{luong2015effective} to calculate the attention energies. As all the models were giving best results with `dot' attention model, we report our final experimental results in the next section only with `dot' attention model.

Table \ref{hyper} shows the detailed hyper-parameter settings used during the evaluation for all three of our architectures. We trained our models on GeForce GTX 1080 GPU with both `Adam' and `SGD' optimizer. All the results in the next section are reported using `SGD' as it was giving comparatively good results. We used PyTorch 0.3.1 for implementing our models under Linux environment.

\begin{table*}[ht]
\centering
\small
\begin{tabular}{|p{2.42in}|l|c|c|c|c|} \hline 
\thead{Model} &  \thead{SE03} &  \thead{nn.} &  \thead{vb.} &  \thead{adj.} &  \thead{adv.} \\ \hline 
Seq2Seq & 66.3 & 47.2 & 45.1 & 59.2 & 68.1 \\ \hline 
Seq2Seq + conv (bigrams) & 59.2 & - & - & - & - \\ \hline 
Seq2Seq + POS (point-wise multiply) & 57.1 & - & - & - & -\\ \hline 
Seq2Seq + POS (weighting) & 67.5 & 49.1 & 58.3 & 61.2 & 68.1\\ \hline 
Seq2Seq + conv + POS (weighting) & \textbf{73.9} & 58.2 & \textbf{70.4} & 71.3 & \textbf{85.4}\\ \hline \hline

\cite{raganato2017neural} BLSTM (att.) &  70.2 & 71.0 & 58.4 & 75.2 & 83.5\\ \hline
\cite{raganato2017neural} Seq2Seq (att.)& 69.6 & 69.5 & 57.2 & 74.5 & 81.8\\ \hline
\cite{raganato2017neural} Seq2Seq (att., LEX, POS) & 68.5 & 70.1 & 55.2 & 75.1 & 84.4\\ \hline \hline
\cite{melamud2016context2vec} Context2Vec & 69.1 & 71.2 & 57.4 & 75.2 & 82.7\\ \hline
\cite{iacobacci2016embeddings} IMS+emb & 70.4 & \textbf{71.9} & 56.6 & \textbf{75.9} & 84.7\\ \hline
\cite{agirre2014random} UKBgloss w2w & 55.4 & 64.9 & 41.4 & 69.5 & 69.7\\ \hline
\cite{moro2014entity} Babelfy & 67.0 & 68.9 & 50.7 & 73.2 & 79.8\\ \hline
\cite{kaageback2016word} BLSTM  & 73.4 & - & - & - & - \\ \hline 
\cite{taghipour2015semi} IMS + adapted CW & 73.4 & - & - & - & - \\ \hline 
\cite{grozea2004finding} Htsa3 &  72.9 & - & - & - & - \\ \hline 
\cite{strapparava2004pattern} IRST - kernels &  72.6 & - & - & - & - \\ \hline 
\cite{lee2004supervised} Nusels &  72.4 & - & - & - & - \\ \hline 

\end{tabular}
\caption{\label{SE}  F1-scores (\%) for English all-words coarse-grained WSD on Senseval 3 dataset }
\end{table*}
\begin{table*}[h]
\small
\centering
\begin{tabular}{|l|c|c|p{0.3in}|p{0.39in}|p{0.39in}|p{0.39in}|p{0.39in}|} \hline
\thead{Model} & \thead{Train} & \thead{Test} & \thead{nn.} & \thead{vb.} & \thead{adj.} & \thead{adv.} & \thead{all} \\ \hline

\multirow{4}{*}{Seq2Seq} & MASC & MASC & 38.2 & 57.1 & 60.8 & 66.5 & 66.4  \\ \cline{2-8}
 & SemCor & SemCor & 36.1 & 45.2 & 41.3 & 49.6 & 70.1  \\ \cline{2-8}
 & SemCor & MASC & 23.2 & 44.1 & 48.1 & 48.9 & 60.3  \\ \cline{2-8}
 & MASC & SemCor & 22.2 & 36.6 & 32.4 & 53.7 & 59.0  \\ \hline \hline
 
\multirow{4}{*}{Seq2Seq + POS (weighting)} & MASC & MASC & 39.9 & 59.7 & 57.5 & 68.8 & 68.1 \\ \cline{2-8}
 & SemCor & SemCor & 37.7 & 47.6 & 41.5 & 50.9 & 76.4  \\ \cline{2-8}
 & SemCor & MASC & 26.6 & 43.3 & 48.3 & 49.0 & 62.6 \\ \cline{2-8}
 & MASC & SemCor & 24.1 & 37.0 & 35.2 & 52.8 & 63.5  \\ \hline \hline

\multirow{4}{*}{Seq2Seq + conv + POS (weighting)} & MASC & MASC & 46.5 & 65.2 & 67.0 & 74.1 & 72.7  \\ \cline{2-8}
 & SemCor & SemCor & 42.1 & 52.1 & 53.9 & 72.5 & 76.3  \\ \cline{2-8}
 & SemCor & MASC & 29.2 & 45.7 & 42.8 & 53.4 & 65.2 \\ \cline{2-8}
 & MASC& SemCor & 26.1 & 40.0 & 38.2 & 53.6 & 65.8  \\ \cline{2-8} \hline
 
\end{tabular}
\caption{\label{pos}  F1-scores (\%) of our top performing models for different parts of speech with different settings of SemCor and MASC corpus }
\end{table*}
\section{Experimental Results}

In this section, we describe in detail our experimental results in terms of F1(\%) score. This section also contains the results of the top performing models for WSD along with the benchmarks. We show how multiple attention can penalize the confusion of the model by allowing it to put more focus on the exact context. Finally, we conclude this section by showing which linguistic feature has more impact on determining a sense of a word. For extensive evaluation, we implemented \texttt{Seq2Seq} with word attention as a baseline model. Also to see the impact of weighted combination of different linguistic features, we implemented \texttt{Seq2Seq + POS (weighting)} model which is similar to \ref{baseline_weighting} except the bigram attention module is being removed.

\begin{table*}[ht]
\small
\centering
\begin{tabular}{|l|p{0.3in}|p{0.39in}|p{0.39in}|p{0.39in}|p{0.39in}|p{0.39in}|} \hline

\thead{Attention weights} &  \thead{Initial value} & \thead{1000th epoch} & \thead{2000th epoch} & \thead{4000th epoch} &\thead{5000th epoch} & \thead{3200th epoch} \\ \hline
On words  $(\ w_1)$ & 0.41 & 0.74 & 0.48 & 0.35 & 0.34 & 0.32 \\ \hline 
On bigrams ${(\ w}_2)$ & 0.33 & 2.02 & 2.69 & 3.12 & 3.65 & 3.45 \\ \hline 
On POS ${(\ w}_3)$ & 0.20 & 2.14 & 2.23 & 3.38 & 3.37 & 3.58 \\ \hline 
\end{tabular}
\caption{\label{weights}  Change in different attention weights with epochs for Seq2Seq + conv + POS (weighting). }
\end{table*}
\begin{figure*}[h]
  \includegraphics[width=\textwidth,height= 5 cm]{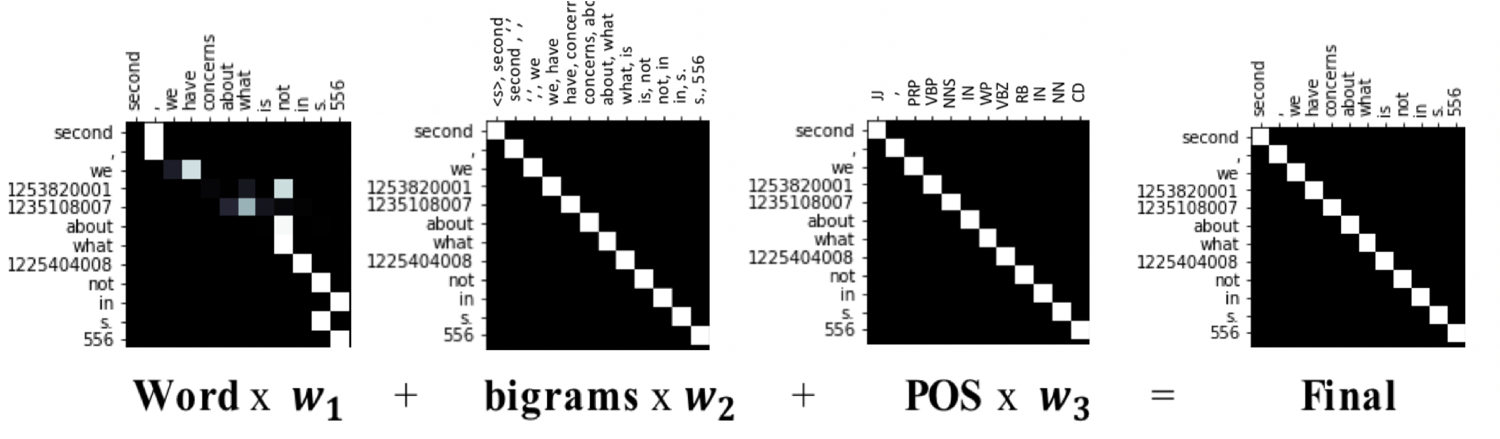}
  \caption{\label{attentions}Effect of different attention weight on final attention matrix. (Numbers on the left of each figure indicates a unique sense id of the corresponding true word on that position)}
\end{figure*}
Table \ref{SE} compares the F1 score achieved by our proposed models against some of the existing state of the art ones on Senseval $3$ task. Our \texttt{Seq2Seq +  conv + POS (weighting)} is the best performing among the five models that we experimented with. It outperformed the top performing neural sequence model from \cite{yuan2016semi} on Senseval-3 task and achieves state of the art F1 score of \textbf{73.9\%}. An interesting aspect is that when POS feature is added through point-wise multiplication, the performance drops (from \textbf{66.3\%} to \textbf{57.1\%}) because it causes inconsistent scaling of different dimension of attention vectors. However, adding the same feature through weighted addition causes a performance boost (from \textbf{66.3\%} to \textbf{67.5\%}) as each dimension of the final attention vectors now gets stretched uniformly based on the two individual attention vectors. Table \ref{SE} also reports the F1 scores of individual POS classes for the Senseval 3 task. Our best model achieves state of the art results on \texttt{$vb.$} and \texttt{$adv.$} classes with F1 scores of \textbf{70.4\%} and \textbf{85.4\%}, respectively. The best results on \texttt{$nn.$} (\textbf{71.9\%}) and \texttt{$adj.$} (\textbf{75.9\%}) are achieved with an IMS framework along with word embeddings to generate features and a support vector machine (SVM) for doing the classification
\cite{iacobacci2016embeddings}. Apart from that, tt can easily be seen that the Seq2Seq architectures perform very well against the statistical and knowledge-based methods like IMS + adapted CW \cite{taghipour2015semi}, Htsa3 \cite{grozea2004finding}, UKBgloss w2w \cite{agirre2014random}, Babelfy \cite{moro2014entity}
as well as RST - kernels \cite{strapparava2004pattern} achieving results that are superior or equivalent to the best models as mentioned above. One interesting evaluation is that the \texttt{Seq2Seq} baseline from \cite{raganato2017neural} is \textbf{69.6\%} and our \texttt{Seq2Seq} baseline performance is \textbf{66.3\%}. When \cite{raganato2017neural} added POS, where it is meant to be learned as one of their tasks, their performance degrades from \textbf{69.6\%} to \textbf{68.5\%}. When we added POS as a feature, our performance jumped from \textbf{66.3\%} to \textbf{67.5\%} which clearly shows that adding POS information as a feature has an influence on identifying polysemy of a word. It is to be noted that, the variation in baseline model performance is may be due to different hyper-parameter settings or different hardware configuration.

Table \ref{pos} shows extensive evaluation of our models on the SemCor and MASC corpora with various training and testing environments. It also shows our model performance on different parts of speech classes on these two corpora. It is clear that whenever testing on the same domain (Train-MASC, Test-MASC and vice versa), all the models perform quite well with maximum F1-score up to \textbf{72.7\%} and \textbf{76.4\%} for MASC and SemCor, respectively. However, while testing on different domains (Train-MASC, Test-SemCor and vice versa), performance of all the  models decreases. It does make sense because even though we are in different domains, we are not tuning any of the hyper-parameters; instead it's been set according to their prior training environment. Table \ref{pos} also depicts how well our best models perform on some frequent parts of speech classes. For almost all of the POS classes, our top performing model is \texttt{Seq2Seq + conv + POS (weighting)}, beating the other models by quite a good margin. Variance in the performance on different parts of speech is mainly because there are some statistically significant differences between the models. However, one thing for sure is that the results for training and testing in different domain is very much correlated with what we have seen in Table \ref{SE}. Also, we have not included the results for \texttt{Seq2Seq + conv (bigrams)} and \texttt{Seq2Seq + POS (point-wise multiply)} as they are comparatively weak models.

Table \ref{weights} depicts the pattern of change in scalar weights on different attentions with epochs which shows how the model decides the amount of attention it needs to pay on different linguistic features. We start with random scalar weights on three possible features as shown in the first column of Table \ref{weights}. It can be easily seen that, with more data the model sees, it gives more more importance to POS with weight \textbf{3.58} compared to the other two (\textbf{0.32} for words and \textbf{3.45} for bigrams). The weight on word attention is not stable but weights on the other two are increasing monotonically until the loss gets very small. Finally, we select the set of weights with which the model achieves highest F1 score at the $3200$th epoch. 

Figure \ref{attentions} shows how the model gives different attentions on different linguistic features. It is clearly visible that with just word attentions, the model gets confused and gives attention to more than one word at a given time. This makes sense for a translation model as one particular word in one language at a specific index position can depend on more than one word of another language. But as we are dealing with same languages on both ends, the decoded word attention has to be on the same word from the encoder; in other words, it should be a one to one mapping. The deviation is mainly because of the lack of context and by making a new attention matrix with the linear weighted combination of multiple attentions on different linguistic features, we penalizes this context lacking. However, this linearly combined attention matrix is finally passed through a \texttt{softmax} layer to make each attention weight a probability.

\section{Conclusion}
In this paper, we adopted a new approach for doing WSD using neural sequence models by applying multiple attentions on different linguistic features of a sentence. The single attention approach with sequence models is very effective with machine translation however this study focuses on using multiple attentions and taking their linear weighted combinations. By making these weights a network parameter, the model can easily fit itself to a suitable combination of them. Our best model achieves state of the art result on Senseval 3 corpus. Also our in depth analysis on POS classes of all the corpora gives us an insight about how polysemy relates with POS. The multiple attention approach has a huge impact on penalizing the mistake made by the model as it gives it more flexibility to choose the right combination from a number of suitable features. This approach can easily be applied to other applications of neural sequence models such as question answering where one can pick the most related fact though an attention over all the fact sentences and finally the answer gets predicted through another attention on question. As future research, we are currently working on this idea.

\bibliographystyle{IEEEtran}
\bibliography{reference}
\end{document}